# Robust SAR ATR on MSTAR with Deep Learning Models trained on Full Synthetic MOCEM data

Benjamin CAMUS
Scalian DS
2 Rue Antoine Becquerel,
35700 Rennes, France.
benjamin.camus@scalian.com

Corentin LE BARBU
Scalian DS
2 Rue Antoine Becquerel,
35700 Rennes, France.
corentin.lebarbu@scalian.com

Eric MONTEUX
Scalian DS
2 Rue Antoine Becquerel,
35700 Rennes, France.
eric.monteux@scalian.com

*Abstract*— the promising potential of Deep Learning for Automatic Target Recognition (ATR) on Synthetic Aperture Radar (SAR) images vanishes when considering the complexity of collecting training datasets measurements. Simulation can overcome this issue by producing synthetic training datasets. However, because of the limited representativeness of simulation, models trained in a classical way with synthetic images have limited generalization abilities when dealing with real measurement at test time. Previous works identified a set of equally promising deep-learning algorithms to tackle this issue. However, these approaches have been evaluated in a very favorable scenario with a synthetic training dataset that overfits the ground truth of the measured test data. In this work, we study the ATR problem outside of this ideal condition, which is unlikely to occur in real operational contexts. Our contribution is threefold. (1) Using the MOCEM simulator (developed by SCALIAN DS for the French MoD/DGA), we produce a synthetic MSTAR training dataset that differs significantly from the real measurements. (2) We experimentally demonstrate the limits of the state-of-the-art. (3) We show that domain randomization techniques and adversarial training can be combined to overcome this issue. We demonstrate that this approach is more robust than the state-of-the-art, with an accuracy of 75 %, while having a limited impact on computing performance during training.

## I. Introduction

The Automatic Target Recognition (ATR) task on Synthetic Aperture Radar (SAR) images consists of designing a classification model to identify objects such as vehicles, captured by a radar. SAR images are (mathematically) complex data characterized by a large dynamic range, an important level of noise (clutter), and a limited resolution. Moreover, they capture multiple and heterogeneous electro-magnetic (EM) effects. Because of these factors, ATR on SAR images is a non-trivial task [1].

Deep Learning algorithms have demonstrated an important potential to tackle this challenge [2, 3, 4]. However, to be efficient these techniques require large training datasets. The problem is then that collecting and labelling a sufficient amount of SAR images is very expensive and in fact impossible in a defense context because, for instance, the vehicles to classify (e.g. tanks, ships) and / or the radar system (deployed on a moving platform like an aircraft or a satellite) may not be available. Moreover, the measurements may contains bias that may fool the models (e.g., classification based on environment features like bushes, instead of targets features) [5].

At the opposite, SAR simulators require only 3D CAD models of the vehicles, materials description and the radar system parameters to generate synthetic images. Besides, a full and strict control of the simulated environment reduces measurement bias, and automates the image labelling process.

For all these reasons, it is interesting to train ATR models using simulated SAR images instead of real measurements. However, because simulations rely on assumptions and real world simplifications (i.e. a model), synthetic data have a limited representativeness. Thus, the synthetic training distribution and the real measurement test distribution differ. This corresponds to the well-known dataset-shift problem [6]: ATR models trained on simulated data have a limited generalization capability to real measurements at test time.

Previous state-of-the-art works [31], especially [7], identified a set of equally promising deep-learning algorithms to tackle this issue. These studies are crucial because they showed that it is theoretically possible to train ATR models with synthetic data and compete with models trained directly on real measurements.

However, the evaluation of these works relies on the MSTAR/SAMPLE [10, 12] datasets with simulated data that reproduce as close as possible the real test measurement. As we show in this paper, the problem is that the vehicles, their configurations (e.g. articulated parts orientations), their equipment, and their positions on the images are almost identical between the synthetic training dataset and the real measurements test dataset, due to a perfect knowledge of ground truth by the authors. That is why, in this work we study the ATR problem outside of this very favorable scenario, unlikely to occur in a real operational context. Our contribution is threefold:

1. Using the MOCEM simulator, we produce a new synthetic training dataset for MSTAR that differs significantly from the real measurements. Thus, these simulated data do not suffer from the aforementioned limitations of the SAMPLE dataset.
2. We experimentally demonstrate the limits of the state-of-art approaches using our synthetic data. We show here that, among all the approaches identified in [7], the Adversarial Training (AT) [8] is the only one that offers acceptable results outside of the ideal conditions of the SAMPLE data. Still, the models accuracy significantly decrease of almost 35 %.

3. We show that intensive domain randomization [9] performed through data augmentation techniques can be combined with AT to overcome this issue and increases the accuracy up to 75 %.

The rest of the paper is as follows. Section II, details the related works and the available SAR datasets. Section III presents our synthetic MOCEM dataset. In Section IV, we demonstrate the limits of the state of the art. Section V presents our approach. Finally, in Section VI we detail our results on the MSTAR dataset and compare our approach to the other algorithms of the state of the art.

## II. STATE OF THE ART

### A. Available datasets

The MSTAR (Moving and Stationary Target Acquisition and Recognition) public dataset [10] comprises SAR measurements of fifteen different targets taken at different depression and azimuth angles by an airborne radar. Following the standard ATR evaluation procedure with MSTAR [11], the 3671 images collected at a depression angle of 17° constitute the training set whereas the 3203 images with a depression angle of 15° serve to test the models. These data concerns ten classes of vehicles (labelled 2S1, BMP2, BDRM2, BTR60, BTR70, D7, T62, T72, ZIL131 and ZSU23-4), measured almost at each azimuth degree from 0° to 360°. Three variants of the vehicles are available for two classes: the BMP2 and the T72. The vehicles equipment and configuration (e.g. side skirts) may differ from a variant to another (see Figure *1*). It is worth noting that the azimuth angles are almost all identical between training and test sets. Moreover, these two sets comprise exactly the same vehicles on same configuration, and they are located at the same positions in the environment. Thus, the model may exploit background information to classify the targets, as the background remains the same between training and test [5].

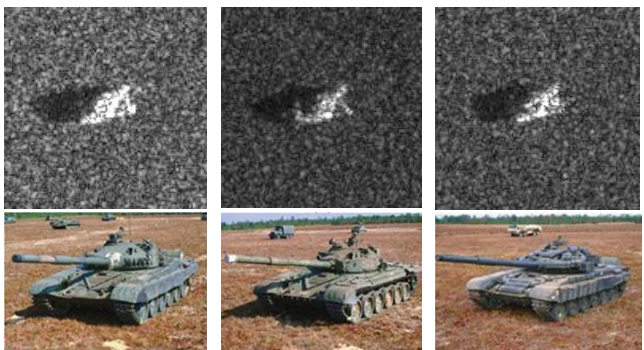

Figure 1 Comparison between SAR images (top) and photo (bottom) of the three MSTAR variants of the T72.

The SAMPLE (Synthetic and Measured Paired Labeled Experiment) public dataset [12] comprises pairs of real SAR measurements and simulated images. This dataset is smaller than MSTAR with only 806 synth-real measurements pairs for training (at depression angles of 14°, 15° and 16°) and 539 pairs for testing (at depression angles of 17°). Like MSTAR, SAMPLE comprises ten target classes labelled 2S1, BMP2, BTR70, M1, M2, M35, M60, M548, T72, and ZSU23-4. One can note that five classes are common with MSTAR. In fact, the real measurements of these five classes correspond to the MSTAR ones (but restricted to azimuth range of SAMPLE). However, SAMPLE data does not provide any variant for any class. The SAMPLE dataset suffers from several drawbacks. First, the images azimuth angles range only from 10° to 80° for both training and test datasets. It particularly excludes the cardinal directions that may be the more challenging angles. Secondly, this angular sector may not be representative of the challenges met when classifying images at a full 360° extent. In addition, the SAMPLE authors have deployed considerable efforts to make the synthetic data as close as possible to real measurements, using detailed ground truth information. Thus, the SAMPLE simulations rely on the CAD models of the actual vehicles used during the MSTAR data collection. Furthermore, the authors configured and enriched the CAD models with detailed based on numerous notes and photos of the MSTAR measurements campaign. This very favorable scenario is unlikely to occur in an actual operational context where the measured vehicles may differ significantly from the CAD models used during training. Finally, the SAMPLE authors performed circular shifts on the measured images to center the targets exactly like in the synthetic images (see an example in Figure *2*). This alignment is unrealistic because it requires a pair of real and synthetic images of the same target captured at the same angle. Hence, it assumes to know the target class a priori, which is not compatible with the ATR problem. It is important to note that absolute target centering (i.e. without pairs of images) is not trivial on SAR images because the vehicles are only partially visible. Thus, the synthetic and measured targets are difficult to align without paired data.

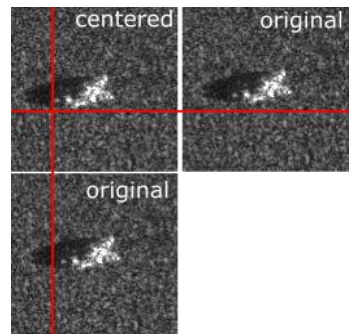

Figure 2. Comparison between an original SAMPLE image (duplicated for comparison purpose) of the T72 and its centered version.

To sum up, MSTAR and SAMPLE are the most complete ATR datasets publicly available in the state-of-the-art. They constitute key contributions to the ATR problem. However, they comprise several biases that may skew experimental results. These two datasets should thus be used with caution.

### B. Related works

Ødegaard et al. got mixed results when training an off-the-shelf deep-learning algorithm directly using simulated SAR data [13]. Using a transfer learning strategy, Malmgren-Hansen et al. pre-trained an ATR classifier with a large amount of synthetic data before training the model with a smaller set of measured images [14]. The drawback of this approach is that it still requires measured data for training, and in many cases, measured images of the target of interest will not be available.

Several works focus on learning a transport function to refine synthetic data by adding features peculiar to measured images. The goal is to transport the synthetic distribution on the measured one to fix the dataset-shift issue. The refined synthetic dataset can then be used to train a regular classifier. Cha et al. trained a residual network to refine synthetic data for ATR [15]. However, their classifier only achieve an accuracy of 55 % at test time. Lewis et al. [16] and Camus et al. [17] trained a GAN to refine synthetic SAR images, with promising results of almost 95 %. However, all these refining approaches require real measurements that may be impossible to obtain. Moreover, Camus et al. demonstrated that GAN are not able to refine new classes never seen during training [17].

Inkawhich et al. trained ATR classifiers with the public synthetic data of SAMPLE by combining several algorithms of the literature designed to improve the generalization of deep-learning models [7]. They evaluate all their models on the SAMPLE measured images. The authors compared three architectures: the network of [18], a ResNet18 [19], and a Wide-ResNet18 [20]. The techniques they studied are the following. Gaussian noise is a form of data augmentation that consists of adding random noise to the training images at runtime [21]. With dropout [22], a random noise is injected during training in the hidden units of the network to erase a given ratio of the detected features. Label smoothing [23] prevents models overconfidence and overfitting by modifying the one-hot labels of the training data so that the correct class probability become smaller than one. Mixup [24] randomly performs convex combinations of pairs of training data and labels. Thus, it promotes linear model behaviors between the training samples. Using a cosine loss [25] instead of the usual cross-entropy, the model is trained to maximize the cosine similarity between its output and the true labels. AT [8] consists of perturbing the training data on a per-pixel basis in order to maximize the model cost (using the gradient information of the network). Thus, the model becomes more robust to adversarial attacks. Finally, the bagging technique [26] averages the prediction of several models that were independently trained. For further details on these strategies, we invite the reader to refer to [7]. With a ResNet18 and several combinations of these techniques (bagging, Gaussian noise and dropout combined either with label-smoothing, mixup, cosine loss or AT), the authors found accuracies of almost 95 %. This work is groundbreaking because it demonstrates the possibility to train ATR models with full synthetic SAR images instead of real measurements. However, the authors used the SAMPLE dataset that contains several flaws, as stated on the previous section. Therefore, the proposed approaches may not be applicable in an actual operational scenario. That is why in the following section we confront the Inkawhich et al. approaches to other, more representative simulated data.

III. SYNTHETIC DATASET PRODUCTION WITH MOCEM

A. Modeling and simulation

To generate our synthetic database for the MSTAR test data, we use the MOCEM software, which is a CAD-based SAR imaging simulator developed by SCALIAN DS for the French MoD (DGA) for 20 years [27, 28]. To be as representative as possible of an operational scenario, we consider three ground-truth fidelity levels in our simulation.
1. What is known with certainty, and should be therefore accurately simulated: the radar sensor
2. What is partially known, and should therefore only be approximated in the simulation: the target signatures using the appropriate CAD models.
3. What is unknown and should thus not be faithfully modeled: the environment of the targets.

We take off-the-shelf CAD models available on Internet. We use one CAD model per MSTAR class. Thus, we do not simulate the variant vehicles for the T72 and the BMP2 classes. We manually simplify the CAD models meshes (when appropriate) to speed-up the simulation time, and we associate electro-magnetic generic materials (i.e. with well-defined reflectivity, roughness and dielectric constant) to the different facets of the model. We configure MOCEM to simulate the transfer function of the MSTAR sensor (i.e. with similar range/cross-range sampling and resolution, thermal noise level, and Taylor window function) (see Figure *3*). We run parametric simulations for the 16°, 17° and 18° incidence angles. For each incidence, we generate images at every 0.5° azimuth for the full [0°, 360°[ range. Thus, we do not consider exactly the same azimuth angles than MSTAR, and our training dataset has different incidence angles than the 15° MSTAR test data. The whole simulation takes two weeks to complete using two standard personal computers.

B. Qualitative evaluation of our dataset

Except for few optical photos per vehicles, we do not know the ground truth of the MSTAR measurements. It is therefore very unlikely that our off-the-shelf CAD models faithfully represent the actual vehicles used during the MSTAR measurement campaign. Moreover, an analysis of the MSTAR photo and SAR images reveals the following differences with ground truth (note that, at the opposite of classical optical images, a small difference on the vehicle geometry may have an important impact on the radar signature of the target).

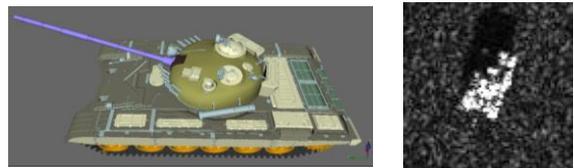

Figure 3. CAD model (left) and synthetic image example (right) of the T72.

The rear spoilers of the BTR60 are too high on the CAD model. Therefore, we observe dihedral effects on our synthetic data that are not present on the measured images. The D7 blade lies on the ground on the MSTAR photo whereas it does not in our simulation (see Figure *4*). Moreover, the dimensions of the CAD model are 25 cm too large, 30 cm too long, and 4 cm too high, according to the D7 user's manual [29]. The fuel barrel of the T62 is too high on our CAD models, and the turret shape does not match the MSTAR vehicle one. Depending on the T72 variant, the CAD turret and the gun orientations are different from ground truth (see Figure *5*); the fuel barrel is missing and the equipment are not the same. The CAD model of the ZIL131 is 15 cm too large, 40 cm too long and 6 cm too high. Moreover, the axletree, the rear bumper, and the Front winch are missing in the CAD model. Finally, the turret orientation of the ZSU23-4 differs between the simulation and the measurements (see Figure 6). The log attached at the back of the ZSU23 is also mistakenly associated with a metal EM material.

The partial knowledge of ground truth and the lack of representativeness of CAD models (especially for equipment and articulated parts configurations) are a guaranty to ensure that our synthetic dataset differs significantly from the real measurements. Hence, it is more representative of the ATR challenges than the SAMPLE data.

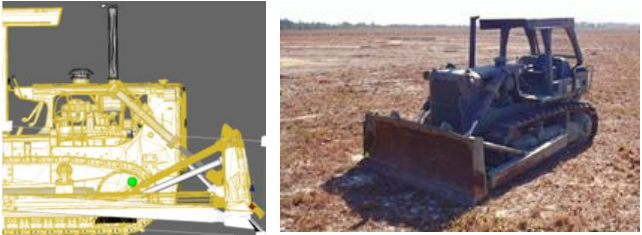

Figure 4. Comparison of our CAD (left) and ground truth (right). The actual blade position is represented in yellow, and the estimated ground truth is in white.

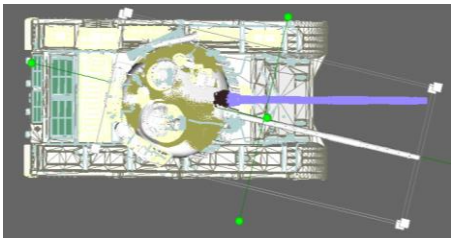

Figure 5. Actual turret position in our T72 CAD model (in violet) versus estimated position (in white).

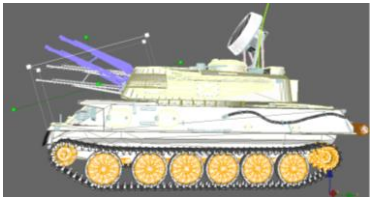

Figure 6. Actual gun position in our ZSU23-4 CAD model (in violet) versus estimated position (in white).

IV. DEMONSTRATION OF THE STATE OF THE ART LIMITS

A. Demonstration using our MOCEM dataset

*1) Datasets:* To study the limits of the state of the art, we apply the algorithms identified by Inkawhich et al. [7] on our MOCEM synthetic data. We evaluate our ATR models with the MSTAR test data (at a depression of 15°). Thus, on the contrary to the SAMPLE dataset used in the original paper, the data ranges on a full 360° azimuth extent, the targets are not pairwise aligned on the synthetic and measured images, and the CAD models of the vehicles does not match the measurements ground truth. We are then closer to a real operational scenario than with the SAMPLE dataset.

*2) Methods:* We use the ResNet18 architecture that gave the best results in the original paper. We follow an experimental plan similar to Inkawhich et al. First, we evaluate a baseline approach that does not involve any improvement techniques. Then, we measure the accuracy obtained with label smoothing (lblsm), mixup, cosine loss and AT applied separately. We also evaluate the models trained with a combination of dropout and Gaussian noise (gauss). Finally, we use Gaussian noise and dropout with either label smoothing, mixup, cosine loss and AT. As the training algorithms are stochastic, we average the accuracy over 50 different trainings for each experimental configuration.

*3) Results:* The "centered" column of Table I shows the results of these experiments (note that we also present the results obtained by Inkawhich et al. on the original SAMPLE data in the table). We observe an important gap between the results on the SAMPLE data and the MOCEM/MSTAR ones. With our MOCEM data, AT gives the best results with only 27 % of accuracy. Considering that this technique achieves an accuracy of almost 77 % on the SAMPLE data, we have a gap of 50 % between SAMPLE and the MOCEM/MSTAR data. At the opposite, the best algorithm with SAMPLE is a combination of label smoothing, Gaussian noise and dropout that produces models with an accuracy of 91 %. This configuration only reaches 20 % in our more realistic context (i.e. a decrease of 71 %). It is important to note that we observe a good convergence of all the training algorithms. Thus, these poor results cannot be detected during training, which makes any fine-tuning of the hyper-parameters irrelevant.

The absence of (unrealistic) targets alignment in the MOCEM/MSTAR data might explain (at least partially) these poor performances. Thus, we test if a simple data augmentation can fix this issue by making the models less sensitive to targets centering. At runtime during training, we apply circular shifts on the synthetic images. The x and y pixel offsets of each shift are uniformly sampled on the $[-5, 5]$ integer interval. We run again all our experiments with this data augmentation. The Table I shows our results in the "rnd shift" column. We observe that circular shift indeed increases the model accuracy. However, the results are still low compared to the performances of Inkawhich et al. AT is still the best with an accuracy of almost 55 %. However, it is 22 % lower than with the SAMPLE data.

TABLE I
AVERAGE ACCURACY OF THE ALGORITHMS IDENTIFIED BY INKAWHICH ET AL. ON OUR MOCEM/MSTAR DATA.

| algorithms | MOCEM training data | | original results on SAMPLE |
|---|---|---|---|
| | centered | rnd shift | |
| baseline | 23.52% | 43.06% | 63.84% |
| lblsm | 23.95% | 44.37% | 77.49% |
| mixup | 21.26% | 35.97% | 80.10% |
| cosine loss | 22.41% | 43.79% | 75.67% |
| **AT** | **27.20%** | **54.48%** | 76.78% |
| gauss, dropout | 21.00% | 30.01% | 88.28% |
| lblsm, gauss, dropout | 20.34% | 31.42% | **91.09%** |
| mixup, gauss, dropout | 21.42% | 28.57% | 90.31% |
| cosine loss, gauss, dropout | 20.04% | 27.70% | 89.10% |
| AT, gauss, dropout | 21.87% | 36.94% | 89.50% |

This experiment demonstrates that we cannot apply the algorithms identified by Inkawhich et al. on our MOCEM/MSTAR data. We formulate two non-exclusive assumptions to explain this fact:

Hypothesis 1: *The algorithms identified by Inkawhich et al. are not adapted to a real operational scenario where the synthetic data does not exactly match the measurements ground truth (like with our MOCEM/MSTAR data).*

Hypothesis 2: *Our MOCEM data are not accurate enough for training deep-learning algorithms (e.g., our simulations might not take account of some important electromagnetic effects).*

We invalidate hypothesis 2 later in Section VI.A using our deep-learning approach. To prove hypothesis 1, we slightly modify the SAMPLE test data to be more compliant with an actual operational scenario, and test the models of Inkawhich et al. on this new dataset.

### B. Demonstration using a modified SAMPLE dataset

*1) Dataset:* First, we used the original not-shifted test measured images of SAMPLE. Thus, the targets are not exactly aligned between the synthetic and the measured images, as expected in a realistic use case. Then, we add the MSTAR images of the variant vehicles that comes from elevation 17° and belongs to the restricted [10°, 80°] azimuth range of SAMPLE. In this way, for the T72 and BMP2 classes, the CAD models of SAMPLE does not match the ground truth of the measured images. It is important to note that these extra MSTAR images in SAMPLE does not skew the results because they originate from the same measurement campaign, and we use the same Quarter Power Magnitude (QPM) LUT [30] on the images. This new dataset is more complex than the original SAMPLE one, but it still corresponds to a very favorable and unlikely scenario because: (1) the azimuth range of [10°, 80°] is too limited, and (2) for 75 % of the synthetic training data, the CAD models still match the ground truth of the measured images.

*2) Results:* However, as shown on the results of Table II, this is enough to show the limits of the algorithms of the literature. The top accuracy is only of 49 % with the mixup, Gaussian noise, dropout combo. This represent a decrease of around 41 % compared to the original SAMPLE test data. The results are better when we add our random circular shift data augmentation but the top accuracy (from the AT, gauss, dropout combination) only reaches 66 % -i.e. -23 % compared to the original SAMPLE data. These weakened results remain however far above the other works of the literature that train ATR models with full synthetic datasets.

TABLE II
AVERAGE ACCURACY OF THE MODELS OF THE LITERATURE TESTED WITH OUR NEW SAMPLE DATASET.

| algorithms | SAMPLE training data | | original results on SAMPLE |
|---|---|---|---|
| | centered | rnd shift | |
| baseline | 33.55% | 38.53% | 63.84% |
| lblsm | 40.30% | 41.82% | 77.49% |
| mixup | 38.50% | 47.94% | 80.10% |
| cosine_loss | 38.48% | 40.33% | 75.67% |
| AT | 39.71% | 47.81% | 76.78% |
| gauss, dropout | 46.90% | 56.26% | 88.28% |
| lblsm, gauss, dropout | 47.87% | 54.84% | **91.09%** |
| mixup, gauss, dropout | **49.22%** | 47.87% | 90.31% |
| cosine_loss, gauss, dropout | 46.11% | 53.87% | 89.10% |
| AT, gauss, dropout | 47.00% | **66.06%** | 89.50% |

### C. Conclusion on the limits of the state of the art

From this series of experiments, we conclude that the algorithms identified by Inkawhich et al. are not adapted to a real operational scenario where the synthetic data does not exactly match the measurements ground truth. This confirm that the SAMPLE dataset is flawed and should be used with caution. In a consistent way with other works of the literature [7, 31], we note that AT has a strong potential to solve the ATR problem as it gives the best performances. However, this technique is not sufficient by itself. In the following section, we detail how we combine AT with an intensive domain randomization strategy to solve this issue.

## V. OUR APPROACH

### A. Method

In our approach, we make a different implementation of AT than Inkawhich et al. To attack the training images we use the Fast Gradient Sign Method (FGSM) [32] instead of the iterative Projected Gradient Descent (PGD) algorithm. Although PGD is theoretically more optimal than FGSM, we find better results with the later method. This may be surprising, but this is consistent with other works in the literature that found better results with FGSM than with PGD for training models to be robust to adversarial attacks [33]. A positive side effect is that FGSM is much faster than PGD because it only requires one additional forward/back-propagation step instead of 50 steps with the PGD of Inkawhich et al. We also find better results when using a L2 norm to bound the adversarial noise (at a value of two in our experiment) instead of the "infinity norm" used by Inkawhich et al.

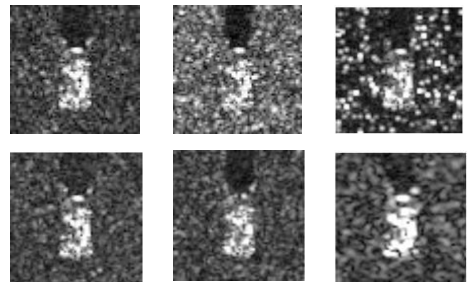

Figure 7. Example of domain randomization: reference image (top left), background clutter level (top center) and distribution (top right), NeSimga0 level (bottom left), target signature (bottom center), and radar resolution (bottom right).

We combine AT with an intensive domain randomization strategy [34]. This technique was originally developed to train robotic systems in a simulated environment. It consists of randomizing the simulated environment parameters to introduce as much variations in the synthetic data as possible. Then, the goal of this heuristic it to stretch the synthetic training distribution so that it eventually cover all the measured test distribution. We adapt this strategy to the SAR ATR problem in the following way. For each training epoch, we create a variant of all our MOCEM dataset by randomly determining for each image separately: the range and cross-range resolution, the level and distribution of the background clutter, the thermal noise of the sensor, and the target position in the images (see Figure 7 for examples). We also select the bright points of the target that are greater than half the maximum amplitude, and randomly dropout (i.e. assign zero values on) half of them to introduce variations in the target signatures. Finally, we use the bagging method to average the prediction of ten independent models.

We also perform Test-Time Data Augmentation (TTDA) at inference-time with our ATR models [37]. We create 20 variants of each test image by applying random circular shifts of [-5, 5] pixels range in the x and y dimensions. Then, we average the model predictions on the 20 variants to classify the image. The Figure 8 sums-up our complete training workflow.

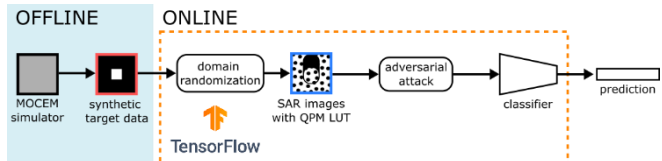

Figure 8. Our complete training workflow.

### B. Architecture, optimizer and hyperparameters

We use a DenseNet121 architecture [38], a SGD (Stochastic Gradient Descent) optimizer with Nesterov momentum [39, 40], a weight decay of $10^{-4}$, and a batch size of 128. The learning rate and the momentum vary during the training according to a "1cycle" policy [41, 42], as shown on Figure 9. We found that AT and the domain randomization slow down the convergence of the training. That is why we train our models for 150 epochs.

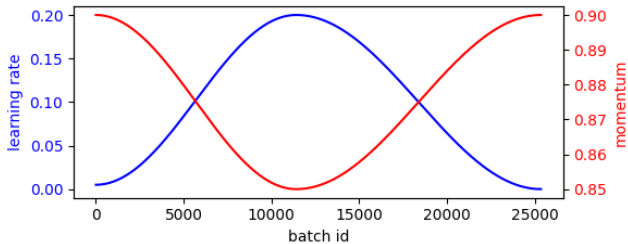

Figure 9. Evolution of the learning rate and momentum in our training according to the "1cycle" policy.

As shown in Table III, we define a uniform distribution for each of the domain randomization variables. We deliberately choose large ranges for the background clutter (level and distribution) and the target positions because this information cannot be known a priori in an operational context [35].

TABLE III
HYPER-PARAMETERS VALUES OF OUR DOMAIN RANDOMIZATION IN OUR EXPERIMENTS.

| parameters | possible value ranges |
|---|---|
| range resolution | [0.203125 m, 0.35 m] |
| cross-range resolution | [0.21 m, 0.35 m] |
| background clutter level | [-20 dB.m²/m², -5 dB.m²/m²] |
| baground clutter gamma distribution parameter | [2., 10.] |
| thermal noise level | [-25 dB.m²/m², -15 dB.m²/m²] |
| circular shift offsets | [-5 px, 5 px] |

To run our experiments, we implement the domain randomization in a Tensorflow (TF) [36] model that performs data augmentation at runtime during the classifier training. This TF model processes batches of target signatures and shadow masks generated with MOCEM. From these inputs, it produces batches of augmented SAR images. Thanks to this TF model, we only have to simulate each training image once. It is important to note that we could achieve the same results by simulating each image with MOCEM specifically for each training epoch with different simulation parameters. Nevertheless, our MOCEM/TF implementation is significantly more efficient. With our TF data-augmentation model, we are indeed able to generate more than 300,000 augmented SAR images per minutes on a single NVIDIA GeForce RTX 2060 GPU. Thus, our domain randomization approach is fast enough to train ATR models in a reasonable time.

## VI. RESULTS

We evaluate our approach and synthetic data through two different experiments. First, we consider a complete and representative use case with the MSTAR test data. Then, to be able to compare our results with the state of the art on SAMPLE, we consider a more restricted scenario.

### A. Evaluating our approach and data with MSTAR

*1) Accuracy of our approach:* Because our training algorithm is stochastic, we train 50 different models with our MOCEM dataset. Then, we test 1e6 combinations of bagging on the MSTAR test data at 15° (including the BMP2 and T72 variants). The average test accuracy of all the bagging is 75.34 %, with a minimum at 74.43 % and a maximum at 76.24 %. Thus, we conclude that with our MOCEM training dataset, our approach gives stable performances that are 48 % higher than the top algorithm of the literature (based on the "centered" results of Table I). This invalidates hypothesis 2, and shows that our MOCEM simulated data are accurate enough to train deep-learning algorithms. Figure 10 shows the confusion matrix of a bagging. Table IV shows the average accuracies per classes over the 1e6 bagging. We observe that for five classes we have an accuracy greater than 80 %. For the two classes with variants –i.e. BMP2 and T72—we have an accuracy greater than 75 %. This shows that we are robust to theses variants.

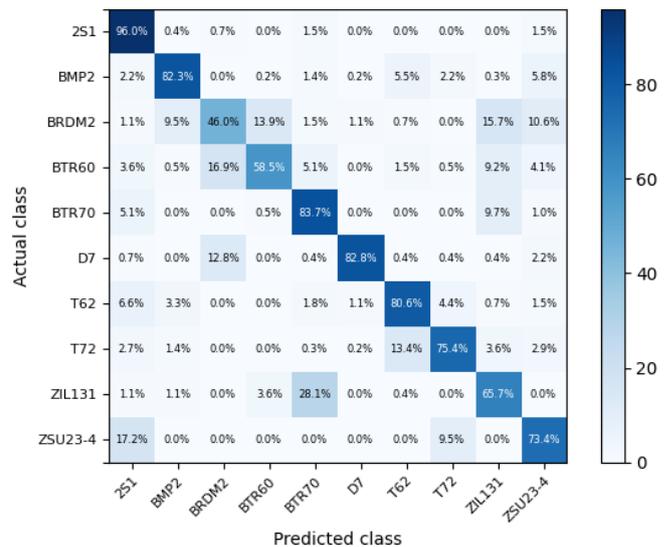

Figure 10. Confusion matrix of a bagging of models.

TABLE IV
AVERAGE ACCURACIES PER CLASSES WITH OUR APPROACH.

|       | mean   | max    | min    |
|-------|--------|--------|--------|
| 2S1   | 95.89% | 96.35% | 94.53% |
| BMP2  | 80.16% | 82.28% | 78.53% |
| BRDM2 | 47.62% | 51.09% | 44.16% |
| BTR60 | 58.99% | 64.10% | 54.36% |
| BTR70 | 83.39% | 84.18% | 81.63% |
| D7    | 81.84% | 85.04% | 78.10% |
| T62   | 80.17% | 82.78% | 77.29% |
| T72   | 77.22% | 79.55% | 74.57% |
| ZIL131| 65.64% | 68.98% | 62.41% |
| ZSU23 | 72.53% | 74.45% | 70.07% |

*2) Ablation study:* We perform an ablation study to measure the contribution of the different techniques composing our approach. The results are shown on Table V. We progressively remove each method and measure the average accuracy on the MSTAR data over 50 different trainings. TTDA and bagging have a negligible impact on the results (of respectively +1.10 % and +0.88 %). Using a DenseNet121 instead of a ResNet18 adds 6.66 % of accuracy. This is because our intensive AT and domain randomization methods makes the function to fit more complex, and therefore requires a bigger architecture, with more capacities [43]. In accordance with our experiments on SAMPLE, we observe that AT significantly increases accuracy of 16.22 %. When removing domain randomization, we need to choose a background clutter level in our synthetic images. This choice is not trivial because it cannot be known a priori [35] and it directly determines the signal-to-noise ratio. We test several levels between -20dB.m²/m² and -5dB.m²/m² with steps of 5dB.m²/m². The worst-case scenario corresponds to a clutter of -20dB.m²/m². Compared to this scenario, domain randomization increases accuracy by 24.95 %. The best-case scenario is with a background clutter of -15 dB.m²/m². This is consistent as it is close to the actual average clutter level of the MSTAR data. Even in this optimal context, domain randomization still brings a gain of 7.42 %. Finally, the random circular shifts increases accuracy of 6.27 % and 19.53 % with respectively the worst and the best background clutter level. To sum-up, AT and domain randomization are two essential components of our approach, as they boost accuracy of about 45 %.

TABLE V
ABLATION STUDY RESULTS

| clutter level without domain rdn (dB.m²/m²) | -20 | -15 |
|---|---|---|
| techniques | accuracy | |
| bagging, DenseNet, rnd shift, domain rnd, AT, TTDA | **75.34%** | |
| bagging, DenseNet, rnd shift, domain rnd, AT | 74.24% | |
| DenseNet, rnd shift, domain rnd, AT | 73.36% | |
| ResNet, rnd shift, domain rnd, AT | 66.70% | |
| ResNet, rnd shift, domain rnd | 50.48% | |
| ResNet, rnd shift | 25.53% | 43.06% |
| ResNet | 19.26% | 23.52% |

### B. Comparing our approach with the state of the art

*1) Methods:* We run the following experiment to compare our synthetic MOCEM dataset and our deep learning approach to the state of the art. On one hand, we train the different algorithms of Inkawhich et al. with the synthetic SAMPLE data. On the other hand, we use our approach to train an ATR model with our MSTAR/MOCEM synthetic data. To be fair, we use the same ResNet18 architecture (and optimizer) than Inkawhich et al. instead of our DenseNet121. We also disable TTDA with our approach.

*2) Datasets:* We test all the models on the same measured dataset without using models bagging. For the approaches to be comparable, we restrict the different datasets to the [10°, 80°] azimuth range of SAMPLE and to the five targets that are common between MSTAR and SAMPLE, namely: 2S1, BMP2, BTR70, T72, and ZSU23-4. To be as representative as possible of an operational context, for testing the models we use our modified SAMPLE measured dataset presented previously in Section IV.B.1 (i.e. without targets centering and with the MSTAR variants for the T72 and the BMP2). It is important to note that this test scenario advantages greatly the state-of-the-art approaches because the CAD models of SAMPLE still (unrealistically) match the measurement ground truth for 60 % of the images. To be fair, we restrict our MOCEM dataset to have the same number of images than SAMPLE (i.e. 386 images) with similar depression and azimuth angles (we select the nearest parametric angles of our dataset).

*3) Results:* Table VI shows the results of this experiment averaged over 50 trainings. We observe that, despite a very unfavorable scenario, our approach has the best results. We are 18 % above the best approach of the literature. When we add a simple random circular shift of [-5, 5] pixel range to the algorithms of Inkawhich et al., we are still 9 % above the top results. With the baseline cases, (i.e. only a ResNet18 without any improvement technique) we observe that the MOCEM and the SAMPLE synthetic datasets give similar results. This may indicate that the two datasets have similar representativeness for SAR ATR. Interestingly, our approach can benefits from additional training images: if we use the all the 2215 images of our MOCEM dataset that belongs to the [10°, 80°] azimuth range, our approach reaches an accuracy of 78.82 % -i.e. +14.74 % compared to a training with only the 386 images that correspond to the SAMPLE angles.

TABLE VI
COMPARISON BETWEEN OUR APPROACH AND THE STATE-OF-THE-ART.

| training data | algorithms | test accuracies | |
|---|---|---|---|
| | | centered training data | training with random shifts |
| SAMPLE | baseline | 35.85% | 49.59% |
| | lblsm | 36.59% | 51.68% |
| | mixup | 32.51% | 45.68% |
| | cosine_loss | 35.42% | 45.75% |
| | AT | 35.32% | 54.62% |
| | gauss, dropout | 43.37% | 42.38% |
| | lblsm, gauss, dropout | 44.01% | 44.73% |
| | mixup, gauss, dropout | 43.83% | 38.36% |
| | cosine_loss, gauss, dropout | 43.58% | 35.69% |
| | AT, gauss, dropout | **45.99%** | **55.48%** |
| MOCEM | baseline | 43.06% | 36.53% |
| | ours | **64.08%** | |

### C. Comparison with models trained on measured data

*1) Motivation:* Our 75 % accuracy on MSTAR may seems low compared to the 95 % achieved by Morgan when

training ATR models directly with the MSTAR measurements [4]. However, as discussed in Section II.A, for this baseline result the models are trained and tested on measured images of exactly the same vehicles, with the same configurations and backgrounds. We run the following experiment to estimate the performance achieved outside of this utopian scenario.

*2) Method and dataset:* We reproduce the Morgan's experiment, with the same algorithm and the MSTAR train and test data of the ten classes. However, we only keep one variant in the training data for the BMP2 and the T72 classes. Then we measure the accuracy of the model on the different variants for these two classes. This way, for two classes the models are tested and trained on different vehicles of the same classes, like in an actual operational scenario. As the training process is stochastic, we average the results over 100 trainings.

*3) Results:* Table VII and VIII shows the results of this experiment. We observe significant accuracy drops when we test the models on new variants never seen at training time. For the T72 class, we have an average accuracy of 93.44 % when the test variant is the same than the train variant. When the variant are different between test and train, the average accuracy falls at 71.59 % (i.e. a decrease of 21.85 %). For the BMP2 class, the average accuracy drops from 80.45 % to 67.65 % (i.e. -12.81 %).

We want to stress that this experiment has a limited scope because it concerns only two classes. However, this may show that the results of the MSTAR literature should be considered cautiously.

On the same two classes, our MOCEM approach has an average accuracy of 80.16 % for the BMP2 and 77.22 % for the T72 (as shown previously on Table IV). Thus, despite using a full synthetic training dataset, we achieve better results with our approach than models trained directly on measured data but with different variants (+12.5 % for the BMP2 and +5.63 % for the T72).

TABLE VII
INFLUENCE OF THE T72 MSTAR VARIANTS (IDENTIFIED BY THEIR SERIAL NUMBER) ON THE ACCURACY.

|  |  | T72 variant used for test | | |
|---|---|---|---|---|
|  |  | 132 | 812 | s7 |
| T72 variant used for training | 132 | 93.13% | 64.78% | 64.95% |
|  | 812 | 72.02% | 96.14% | 72.42% |
|  | s7 | 70.92% | 84.31% | 91.00% |

TABLE VIII
INFLUENCE OF THE BMP2 MSTAR VARIANTS (IDENTIFIED BY THEIR SERIAL NUMBER) ON THE ACCURACY.

|  |  | BMP2 variant used for test | | |
|---|---|---|---|---|
|  |  | 9563 | 9566 | c21 |
| BMP2 variant used for training | 9563 | 77.49% | 60.07% | 60.44% |
|  | 9566 | 72.82% | 83.40% | 60.47% |
|  | c21 | 81.86% | 69.82% | 80.48% |

VII. CONCLUSION

In this work, we study the problem of training ATR classifiers that transfer to real measurements using synthetic SAR images. We demonstrate that the publicly available datasets of the literature (the MSTAR and SAMPLE data) are limited because they are not fully representative of the challenges met in a real operational context. In particular, their training set unrealistically match the ground truth of their measure test data (e.g., the CAD models of SAMPLE are very close to the measured vehicles, the position of the targets are identical on the images). These dataset remains however essential in the literature to conduct experiments. That is why they should be used cautiously.

Considering this potential skew, we show the limitation of the key set of approaches identified by Inkawhich et al. using the SAMPLE dataset. We show that with slight modifications of the test data (removing the targets alignment and adding variants of vehicles for two classes) the top accuracy decreases of 42 % (from 91 % to 49 %).

Using the MOCEM simulator, we produce a new synthetic training dataset for MSTAR that differs significantly from the real measurements. This dataset is then more representative of the ATR challenges. We show that the approaches of the literature give poor results (i.e. only 54 % of top accuracy) with our synthetic dataset. Thanks to all our experiments, we identified that among all the methods proposed by Inkawhich et al., AT is the most promising. However, we conclude that this approach is not sufficient by itself.

That is why we propose a new approach that combines AT with an intensive SAR-based domain randomization technique. Following this strategy, at each epoch we randomly change the training images range and cross-range resolution, the level and distribution of the background clutter, the thermal noise of the sensor, the target position in the images, and the target signature. We implement this domain randomization in a Tensorflow model that performs data augmentation at runtime during training directly on the GPU. However, we could achieve the same results (but less efficiently) by simulating each images with MOCEM specifically for each training epoch with different simulation parameters.

With this approach and our full synthetic MOCEM training data, we achieve an accuracy of 75 % on the MSTAR test data. These results are 48 % above the top results obtained when training the approaches of the literature with our MOCEM dataset. We also show that with MOCEM data, our solution achieves an accuracy 18 % above the top algorithms of literature trained with the SAMPLE synthetic data. In an actual scenario, our solution might give similar results than models trained directly on measured data with classical algorithms, although further experimentations should be done to confirm this point. This work demonstrates that our approach is relevant and that the MOCEM simulator is accurate enough to generate representative synthetic datasets, able to train deep-learning algorithms.

In future works, we plan to extend our domain randomization strategy to modify other SAR-based simulation parameters. To identify these new parameters, we plan to use explicability tools to understand what features are essential to correctly classify the images. We also want to perform a more advanced ablation study to measure which parameters of the domain randomization are essential.

## REFERENCES

[1] L. M. Novak, G. J. Owirka and W. S. Brower, "An efficient multi-target SAR ATR algorithm," Conference Record of ACSSC, 1998, pp. 3-13

[2] A. El Housseini, A. Toumi and A. Khenchaf, "Deep Learning for target recognition from SAR images," 2017 Seminar on DAT, 2017, pp. 1-5.

[3] Y. Li, et al. "DeepSAR-Net: Deep convolutional neural networks for SAR target recognition," 2017 IEEE ICBDA, 2017, pp. 740-743.

[4] D. Morgan. "Deep convolutional neural networks for ATR from SAR imagery." Algorithms for Synthetic Aperture Radar Imagery XXII. Vol. 9475. SPIE, 2015.

[5] C. Belloni. Deep learning and featured-based classification techniques for radar imagery. Computer Vision and Pattern Recognition. PhD thesis of Ecole nationale supérieure Mines-Télécom Atlantique, 2019.

[6] J. Quionero-Candela, M. Sugiyama, A. Schwaighofer, and N.D. Lawrence. 2009. Dataset Shift in Machine Learning. The MIT Press.

[7] N. Inkawhich et al., "Bridging a Gap in SAR-ATR: Training on Fully Synthetic and Testing on Measured Data," in IEEE Journal of Selected Topics in Applied Earth Observations and Remote Sensing, vol. 14, pp. 2942-2955, 2021.

[8] A. Madry, A. Makelov, L. Schmidt, D. Tsipras and A. Vladu, "Towards deep learning models resistant to adversarial attacks", Proc. Int. Conf. Learn. Representations, 2018.

[9] Tobin et al. Domain randomization for transferring deep neural networks from simulation to the real world. IEEE IROS 2017

[10] https://www.sdms.afrl.af.mil/index.php?collection=mstar

[11] T. Ross, S. Worrell, V. Velten, J. Mossing and M. Bryant, "Standard SAR ATR evaluation experiments using the MSTAR public release data set," Proc. SPIE 3370, Algorithms for Synthetic Aperture Radar Imagery V, 1998

[12] B. Lewis, et al. "A SAR dataset for ATR development: the Synthetic and Measured Paired Labeled Experiment (SAMPLE)." Algorithms for Synthetic Aperture Radar Imagery XXVI. Vol. 10987. SPIE, 2019.

[13] N. Ødegaard, A. O. Knapskog, C. Cochin and J. Louvigne, "Classification of ships using real and simulated data in a convolutional neural network," 2016 IEEE RadarConf, 2016.

[14] D. Malmgren-Hansen et al., "Improving SAR Automatic Target Recognition Models With Transfer Learning From Simulated Data," in IEEE GRSL, vol. 14, no. 9, Sept. 2017.

[15] M. Cha, et al., "Improving Sar Automatic Target Recognition Using Simulated Images Under Deep Residual Refinements," IEEE ICASSP, 2018.

[16] B. Lewis, J. Liu, A. Wong, "Generative adversarial networks for SAR image realism," Proc. SPIE, Algorithms for Synthetic Aperture Radar Imagery XXV, 1064709, 2018.

[17] B. Camus, E. Monteux and M. Vermet. Refining Simulated SAR images with conditional GAN to train ATR Algorithms. In Proceedings of the Conference on Artificial Intelligence for Defense (CAID), 2020.

[18] Y. Lecun, L. Bottou, Y. Bengio, and P. Haffner, "Gradient-based learning applied to document recognition," Proc. IEEE, vol. 86, no. 11, pp. 2278–2324, Nov. 1998.

[19] K. He, X. Zhang, S. Ren, and J. Sun, "Deep residual learning for image recognition," in Proc. IEEE Conf. Comput. Vis. Pattern Recognit.Comput. Soc., 2016, pp. 770–778

[20] S. Zagoruyko and N. Komodakis, "Wide residual networks," in Proc. Brit. Mach. Vis. Conf., 2016, pp. 87.1–87.12.

[21] C. Shorten and T. M. Khoshgoftaar, "A survey on image data augmentation for deep learning", J. Big Data, vol. 6, 2019.

[22] N. Srivastava, G. E. Hinton, A. Krizhevsky, I. Sutskever and R. Salakhutdinov, "Dropout: A simple way to prevent neural networks from overfitting", J. Mach. Learn. Res., vol. 15, no. 1, pp. 1929-1958, 2014.

[23] R. Möller, S. Kornblith and G. E. Hinton, "When does label smoothing help", Proc. Adv. Neural Inf. Process. Syst, pp. 4696-4705, 2019.

[24] H. Zhang, M. Cissé, Y. N. Dauphin and D. Lopez-Paz, "Mixup: Beyond empirical risk minimization", Proc. Int. Conf. Learn. Representations, 2018.

[25] B. Barz and J. Denzler, "Deep learning on small datasets without pre-training using cosine loss", Proc. IEEE Winter Conf. Appl. Comput. Vis., pp. 1360-1369, 2020.

[26] I. Goodfellow, Y. Bengio and A. Courville, Deep Learning, 2016, [online] Available: http://www.deeplearningbook.org.

[27] C. Cochin, P. Pouliguen, B. Delahaye, D. Le Hellard, P. Gosselin, and F. Aubineau, "Mocem - an 'all in one' tool to simulate sar image," pp. 1 – 4, 07 2008.

[28] C. Cochin, J.-C. Louvigne, R. Fabbri, C. Le Barbu, A. O Knapskog, and N. Ødegaard, "Radar simulation of ship at sea using mocem v4 and comparison to acquisitions," 10 2014.

[29] Direct Support and General Support Maintenance Manual (Caterpillar Model D7F). Department of the Army Technical Manual, 1971.

[30] Doerry, Armin W. SAR Image Scaling Dynamic Range Radiometric Calibration and Display. United States: N. p., 2019. Web. doi:10.2172/1761879.

[31] B. Lewis, B. K. Cai, and C. Bullard. "Adversarial training on SAR images." Automatic Target Recognition XXX. Vol. 11394. International Society for Optics and Photonics, 2020.

[32] Ian J Goodfellow, Jonathon Shlens, and Christian Szegedy. Explaining and harnessing adversarial examples. arXiv preprint arXiv:1412.6572, 2014.

[33] Wong, E., Rice, L., & Kolter, J. Z. (2020). Fast is better than free: Revisiting adversarial training. arXiv preprint arXiv:2001.03994.

[34] Tobin et al. Domain randomization for transferring deep neural networks from simulation to the real world. IEEE IROS 2017

[35] Ulaby, F., Dobson, M. C., & Álvarez-Pérez, J. L. (2019). Handbook of radar scattering statistics for terrain. Artech House.

[36] M. Abadi, A. Agarwal, P. Barham, et al. TensorFlow: Large-scale machine learning on heterogeneous systems, 2015. Software available from tensorflow.org.

[37] Simonyan, K., & Zisserman, A. (2014). Very deep convolutional networks for large-scale image recognition. arXiv preprint arXiv:1409.1556.

[38] Huang, G., Liu, Z., Van Der Maaten, L., & Weinberger, K. Q. (2017). Densely connected convolutional networks. In Proceedings of the IEEE conference on computer vision and pattern recognition (pp. 4700-4708).

[39] Yurii Nesterov. A method of solving a convex programming problem with convergence rate o (1/k2). In Soviet Mathematics Doklady, volume 27, pages 372–376, 1983

[40] Ilya Sutskever, James Martens, George E Dahl, and Geoffrey E Hinton. On the importance of initialization and momentum in deep learning. ICML (3), 28:1139–1147, 2013.

[41] Smith, L. N., & Topin, N. (2019, May). Super-convergence: Very fast training of neural networks using large learning rates. In Artificial Intelligence and Machine Learning for Multi-Domain Operations Applications (Vol. 11006, p. 1100612). International Society for Optics and Photonics.

[42] Smith, L. N. (2017, March). Cyclical learning rates for training neural networks. In 2017 IEEE winter conference on applications of computer vision (WACV) (pp. 464-472). IEEE.

[43] Xie, C., and Yuille, A. Intriguing properties of adversarial training at scale. In Proceeding ICLR 2020